\documentclass{article}

\usepackage{arxiv}

\usepackage[utf8]{inputenc} 
\usepackage[T1]{fontenc}    
\usepackage{hyperref}       
\usepackage{url}            
\usepackage{booktabs}       
\usepackage{amsfonts}       
\usepackage{nicefrac}       
\usepackage{microtype}      
\usepackage{lipsum}
\usepackage{multirow}
\usepackage{hhline}
\usepackage{graphicx}
\usepackage{amsmath,amssymb}

\usepackage{xcolor}
\definecolor{RoyalBlue}{RGB}{65, 105, 225}

\title{Improved Trainable Calibration Method for Neural Networks on Medical Imaging Classification}

\author{Gongbo~Liang\textsuperscript{1},~~ 
		Yu~Zhang\textsuperscript{1},~~
		Xiaoqin~Wang\textsuperscript{2},
		Nathan~Jacobs\textsuperscript{1} \\ 
		1. Department of Computer Science, University of Kentucky~~\\
		2. Department of Radiology, University of Kentucky \\
		{\tt\small \{gb.liang, y.zhang, xiaoqin.wang, nathan.jacobs\}@uky.edu}
		}
		
\begin{document}
\maketitle

\begin{abstract}
Recent works have shown that deep neural networks can achieve super-human performance in a wide range of image classification tasks in the medical imaging domain. However, these works have primarily focused on classification accuracy, ignoring the important role of uncertainty quantification. Empirically, neural networks are often miscalibrated and overconfident in their predictions. This miscalibration could be problematic in any automatic decision-making system, but we focus on the medical field in which neural network miscalibration has the potential to lead to significant treatment errors. We propose a novel calibration approach that maintains the overall classification accuracy while significantly improving model calibration. The proposed approach is based on expected calibration error, which is a common metric for quantifying miscalibration. Our approach can be easily integrated into any classification task as an auxiliary loss term, thus not requiring an explicit training round for calibration. We show that our approach reduces calibration error significantly across various architectures and datasets. 
\let\thefootnote\relax\footnotetext{This paper is accepted to BMVC2020.}
\let\thefootnote\relax\footnotetext{For more information, please visit our project page:~\href{http://www.gb-liang.com/dca}{www.gb-liang.com/dca}.}
\end{abstract}

\section{Introduction}
Recent advances in deep learning research have dramatically impacted the research field of medical imaging analysis~\cite{sahiner2019deep,suzuki2017overview,liang2019ganai}. Many high-performance deep learning models have been developed in the field~\cite{esteva2017dermatologist,yang2018low,mihail2019automatic,yu20192d}. Researchers are actively pushing convolutional neural networks (CNNs) to have higher and higher accuracy, while uncertainty quantification is often ignored when evaluating these models~\cite{ronneberger2015u,ribli2018detecting,liang2019joint,xing2020dynamic,wang2020inconsistent}. However, uncertainty quantification of neural networks is important, especially in automatic decision-making settings in the medical field. An automated method that achieves high accuracy, but captures uncertainty inaccurately, such as providing inaccurate confidence or probability of a specific prediction, could lead to significant treatment errors~\cite{jiang2011calibrating}.

\begin{figure}[!tb]
    \centering
    \includegraphics[width=0.95\textwidth]{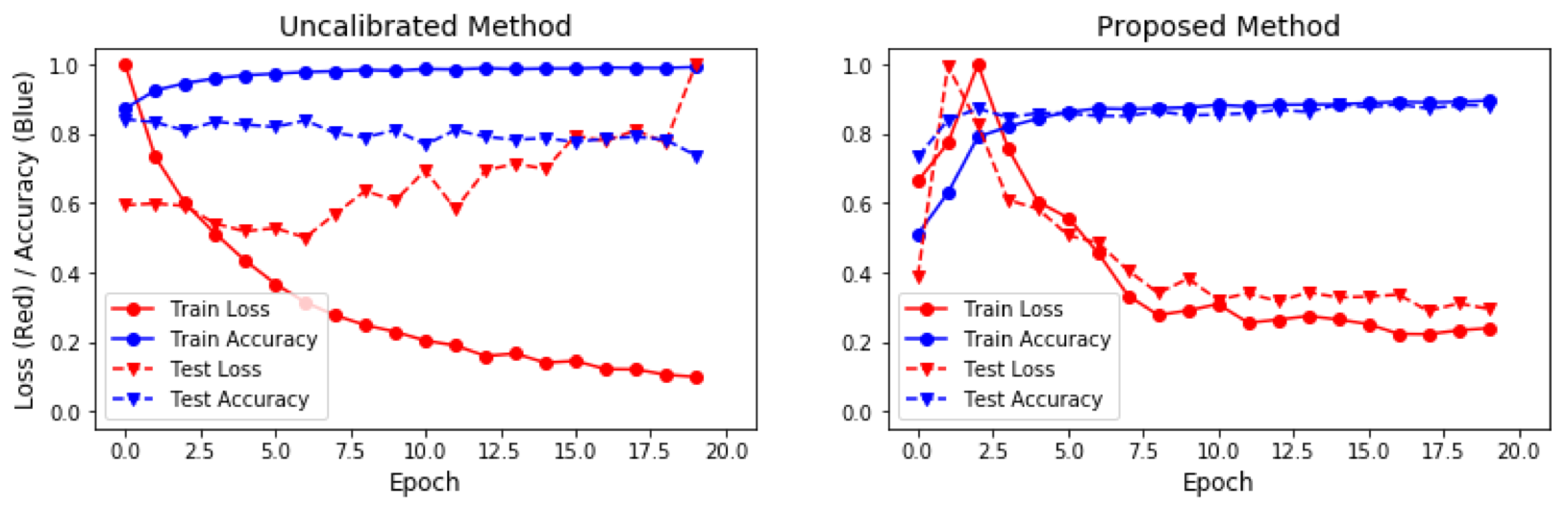}
    \caption{Left: The train loss/accuracy and test loss/accuracy of the uncalibrated model. The model is overfitted after the $7^{th}$ epoch, where the train loss keeps decreasing but the test loss keeps increasing. 
    Right: The train loss/accuracy and test loss/accuracy of our method. The DCA term penalizes the model when the loss reduces but the accuracy is plateaued. Both the train and test losses maintain at the same level after the $7^{th}$ epoch.}
    \label{fig:first_page_figure}
\end{figure}

Unfortunately, deep neural networks are poorly calibrated~\cite{pereyra2017regularizing,kumar2018trainable}, which are overconfident in their predictions~\cite{guo2017calibration,pereyra2017regularizing,kumar2018trainable}. One reason for miscalibration of classification models is that the models can overfit the cross-entropy loss easily without overfitting the $0/1$ loss (i.e., accuracy)~\cite{guo2017calibration,zhang2016understanding}. We propose to add the difference between predicted confidence and accuracy (DCA) as an auxiliary loss for classification model calibration. The DCA term applies a penalty when the cross-entropy loss reduces but the accuracy is plateaued (Figure~\ref{fig:first_page_figure}). 

We evaluate the proposed method across four public medical datasets and four widely used CNN architectures. The results show that our approach reduces calibration error significantly by an average of $65.72\%$ compared to uncalibrated methods (from $0.1006$ ECE to $0.0345$ ECE), while maintaining the overall accuracy across all the experiments---$83.08\%$ and $83.58\%$ for the uncalibrated method and our method, respectively.

\section{Background}
The problem we are addressing is the miscalibration issue of deep neural networks for classification tasks.
The confidence associated with a prediction (i.e., probability of being one specific class) should reflect the true correctness likelihood of a model~\cite{guo2017calibration}. However, deep neural networks tend to be overconfident in their predictions~\cite{pereyra2017regularizing,kumar2018trainable}.

\subsection{Problem Definition}
Mathematically, the problem can be defined in the following way. The input $X \in x$ and label $Y \in y = \{1, ..., k\}$ are random variables that follow a joint distribution $\pi(X,Y) = \pi(Y|X)\pi(X)$. Let $h$ be a deep neural network with $h(X)=(\hat Y,\hat P)$, where $\hat Y$ is the predicted class label and $\hat P$ is the associated confidence. We would like the confidence estimate $\hat P$ to be calibrated, which intuitively means that $\hat P$ represents a true probability. For instance, given $100$ predictions with the average confidence of $0.95$, we expect that $95$ predictions should be correct. In reality, the average confidence of a deep neural network is often higher than its accuracy~\cite{guo2017calibration,pereyra2017regularizing,kumar2018trainable}. The perfect calibration can be defined as: 
\begin{equation}
\mathbb{P}\left(\hat Y = Y | \hat P = p\right) = p, \forall p\in[0,1].
\end{equation}
Difference in expectation between confidence and accuracy (i.e., the calibration error) can be defined as:
\begin{equation}
\label{eq:expectation_dca}
\mathbb{E}_{\hat p}\left[\left| \left(\hat Y = Y | \hat P = p\right) - p\right|\right].
\end{equation}

\subsection{Measurements}
Expected Calibration Error (ECE) is a commonly used criterion for measuring neural network calibration error. ECE~\cite{naeini2015obtaining} approximates Equation~\eqref{eq:expectation_dca} by partitioning predictions into $M$ bins and taking a weighted average of the accuracy/confidence difference for each bin. All the samples need to be grouped into $M$ interval bins according to the predicted probability. Let $B_m$ be the set of indices of samples whose predicted confidence falls into the interval $I_m=(\frac{m-1}{M}, \frac{m}{M}]$, $m \in M$. The accuracy of $B_m$ is
\begin{equation}
\label{eq:acc}
\text{acc}(B_m) = \frac{1}{|B_m|} \sum_{i\in B_m} 1(\hat y_i = y_i), 
\end{equation}
where $\hat y_i$ and $y_i$ are the predicted and ground-truth label for sample $i$.
The average predicted confidence of bin $B_m$ can be defined as 
\begin{equation}
\label{eq:conf}
\text{conf}(B_m) = \frac{1}{|B_m|} \sum_{i\in B_m} \hat p_i, 
\end{equation}
where $\hat p_i$ is the confidence of sample $i$. ECE can be defined with $\text{acc}(B_m)$ and $\text{conf}(B_m)$ 
\begin{equation}
\label{eq:ece}
\text{ECE} = \sum_{m=1}^{M} \frac{|B_m|}{n} \left|\text{acc}(B_m) - \text{conf}(B_m)\right|,
\end{equation}
where $n$ is the number of samples.

Maximum Calibration Error (MCE)~\cite{naeini2015obtaining} is another common criterion for measuring neural network calibration error that partitions predictions into M equally-spaced bins and estimates the worst-case scenario. MCE can be computed as:
\begin{equation}
\label{eq:mce}
\text{MCE} = \max_{m\in \{1,...,m\}}|\text{acc}(B_m) - \text{conf}(B_m)|.
\end{equation}

\section{Existing Calibration Methods}
In this section, we introduce some existing calibration methods, including temperature scaling~\cite{hinton2015distilling,guo2017calibration}, entropy regularization~\cite{pereyra2017regularizing}, MMCE regularization~\cite{kumar2018trainable}, label smoothing~\cite{szegedy2016rethinking,muller2019does}, and Mixup training~\cite{zhang18mixup,thulasidasan2019onmixup}. Temperature scaling is a widely used calibration method, which treats model calibration as a post-processing task. All the other methods fix neural network calibration during the classification training stage.

\subsection{Temperature Scaling}
\label{sec:ts}
Temperature scaling~\cite{hinton2015distilling,guo2017calibration} is a widely-used approach for deep learning model calibration. It fixes the miscalibration issue by dividing the logits by a temperature parameter of $T$ ($T>0$). 
The method involves two steps, in general. The first step is to train a classification model. Once the model is trained, the temperature parameter is added to the model and needs to be trained on the validation set while all the other parameters are frozen~\cite{guo2017calibration}. After that, the temperature parameter will be used for calibration at the testing time. The calibrated confidence, $\hat q_i$, using temperature scaling is
\begin{equation}
\hat q_i = \max_k \theta_{SM}(\frac{z_i}{T})^{(k)},
\end{equation}
where $k$ is the class label ($k = 1,...,K$), $\theta_{SM}(z_i)$ is the predicted confidence. As $T\rightarrow\infty$, the confidence $\hat q_i$ approaches the minimum, which indicates maximum uncertainty. 
 
Temperature scaling is easy to use and performs well. The optimization process of the temperature parameter is not expensive and only needs to be done once. However, as a post-processing approach, temperature scaling does not help with feature learning. In addition, a neural network model should be able to calibrate itself without any post-processing~\cite{widmann2019calibration}. 
 
\subsection{Trainable Calibration Methods}
Trainable calibration methods are proposed to integrate model calibration into classification training. No explicit training round for calibration is needed in such a fashion. One of the earliest trainable approaches is the entropy regularization~\cite{pereyra2017regularizing}. The method proposes to use entropy as a regularization term in loss functions for model calibration. The final classification loss can be written as: 
\begin{equation}
\label{eq:mmcd}
\begin{split}
\text{Loss} &= \text{CrossEntropy} + \beta \text{Entropy},
\end{split}
\end{equation}
where $\beta$ is a weight scalar. One disadvantage of entropy regularization is that the method is very sensitive to the value of $\beta$~\cite{kumar2018trainable}.
Kumar et al. propose to use MMCE replacing entropy for model calibration~\cite{kumar2018trainable}.  MMCE is computed in a reproducing kernel Hilbert space (RKHS)~\cite{gretton2013introduction}. 
The completely loss function can be written as:
\begin{equation}
\label{eq:mmcd}
\begin{split}
\text{Loss} &= \text{CrossEntropy} + \beta (\text{MMCE}^2_m(D))^{\frac{1}{2}},
\end{split}
\end{equation}
where $D$ denotes a dataset. The performance of MMCE may be limited by imbalance predictions of a neural network. For instance, the number of correct predictions is usually larger than the number of incorrect predictions. Thus, the MMCE term needs to be re-weighted as:
\begin{equation}
\label{eq:mmcd_w}
\begin{split}
\text{MMCE}^2_w &= \sum_{c_i=c_j=0} \frac{p_i, p_j, k(p_i, p_j)}{(m-n)^2} +\\ 
& \sum_{c_i=c_j=1} \frac{(1-p_i)(1-p_j)k(p_i, p_j)}{n^2} - \\
& 2 \sum_{c_i=1, c_j=0} \frac{(1-p_i)p_jk(p_i, p_j)}{(m-n)n},
\end{split}
\end{equation}
where $c$ is the predicted label, $m$ is the number of correct predictions, $n$ is the batch size, and $k$ is a universal kernel~\cite{song2008learning}.

Label smoothing~\cite{szegedy2016rethinking} was proposed to improve classification performance of the Inception architecture. 
Müller et al. demonstrate that label smoothing improves classification performance by calibrating models implicitly~\cite{muller2019does}. Instead of targeting a hard probability, $1.0$, for the correct class, label smoothing tries to predict a softer version of it: 
\begin{equation}
\label{eq:mmcd}
\begin{split}
y^{LS}_k = y_k(1-\alpha)+\frac{\alpha}{K},
\end{split}
\end{equation}
where $y_k$ is original targeting probability ($y_k=1.0$ for the correct class and $y_k = 0.0$ for the rest), $K$ is the number of class labels, $\alpha$ is a hyperparameter that determines the amount of smoothing. Mixup~\cite{zhang18mixup} is another method that aims to predict a softer target by randomly mixing training samples. During the training, two samples from different classes are randomly mixed together. Instead of predicting one target label, the network needs to predict the two corresponding labels' probability. The target probabilities equal to the portion of the pixels from each image. Thulasidasan et al.~\cite{thulasidasan2019onmixup} demonstrate that Mixup is also useful for neural network calibration. 

\section{Proposed Method}
\label{sec:dca}
We propose to add the difference between confidence and accuracy (DCA) as an auxiliary loss term to the cross-entropy loss for classification tasks. DCA is based on expected calibration error by minimizing the difference between the predicted confidence and accuracy directly. Therefore, the proposed method can calibrate neural networks effectively. The proposed method is easy to implement and suitable for any classification tasks. In general, classification loss can be written as follows:
\begin{equation}
\text{Loss} = -\frac{1}{N}\sum_{i=1}^N y_i \cdot \log(p(y_i))+\beta DCA,
\end{equation}
where $y_i$ is the true label and $p(y_i)$ is the predicted confidence (i.e., probability) of the true label. The DCA term can be computed for each mini-batch using the following equation: 
\begin{equation}
\label{eq:dca}
\begin{split}
\text{DCA} &= \left|\frac{1}{N}\sum_{i=1}^N c_i - \frac{1}{N}\sum_{i=1}^N p(\hat y_i)\right|,
\end{split}
\end{equation}
where $\hat y_i$ is the predicted label; $c_i=1$, if $\hat y_i = y_i$; otherwise, $c_i=0$. The final loss function can be written as:
\begin{equation}
\text{Loss} = -\frac{1}{N}\sum_{i=1}^N y_i \cdot \log(p(y_i))+\beta \left|\frac{1}{N}\sum_{i=1}^N c_i - \frac{1}{N}\sum_{i=1}^N p(\hat y_i)\right|.
\end{equation}

The DCA auxiliary loss fixes the miscalibration issue by penalizing deep learning models when the cross-entropy loss can be reduced, but the accuracy does not change (i.e., when the model is overfitting). The term forces the average predicted confidence to match the accuracy over all training examples without strict constraint on each example, which pushes the network closer to the ideal situation, in which the accuracy reflects the true correctness likelihood of a model. The averaging mechanism of DCA also smooths the predictions that have extremely high or low confidence. 

DCA is differentiable in the predicted confidence term but not strictly in the prediction accuracy term due to the argmax step for computing the predicted label. During the training phase, gradients can be backpropagated through the confidence terms but not through the accuracy.

\section{Experiments}
We compare the proposed method with temperature scaling and uncalibrated models (trained with cross-entropy loss without applying any calibration methods) on four medical imaging datasets across four popular CNN networks. The trainable methods are not compared in this work because the literature shows that they have a worse or similar calibration performance with temperature scaling~\cite{kumar2018trainable,muller2019does,thulasidasan2019onmixup}. Thus, it may not be necessary to be compared in this paper explicitly.

The evaluation results show that the proposed method significantly improves model calibration while maintaining the overall classification accuracy. The proposed method reduces calibration error by an average of $65.72\%$ compared to uncalibrated methods (from $0.1006$ ECE to $0.0345$ ECE) and performs about $20\%$ better than temperature scaling on average. 

\subsection{Experiment Setup}
Four medical imaging datasets (RSNA~\cite{RSNA2019kaggle}, DDSM~\cite{heath2000digital}, Mendeley V2~\cite{kermany2018labeled}, and Kather 5000~\cite{kather2016multi}) were used in this study for both binary and multi-class classification tasks. See Table~\ref{table:datasets} and Section~\ref{sec:s2} for more details. Four CNN models that trained with transfer learning mechanism were evaluated. More specifically, we use ImageNet~\cite{deng2009imagenet} pre-trained AlexNet~\cite{krizhevsky2012imagenet}, ResNet-50~\cite{he2016deep}, DenseNet-121~\cite{huang2017densely}, and SqueezeNet 1-1~\cite{iandola2016squeezenet} as fixed feature extractors, a $1\times 1$ convolutional layer and two fully connected layers were added on top to each feature extractor. See Section~\ref{sec:s3} for more details.

\subsection{Calibration Results}
Table~\ref{table:ece_result} shows the expected calibration error (ECE) and the accuracy of the uncalibrated models (Unca.), temperature scaling (Temp.), and the proposed method (DCA). Each model was trained for two times. The average value is shown in the table. 

The table shows that our method is consistently better than the uncalibrated method on model calibration, which reduces the ECE by $65.71\%$ on average (from $0.1006$ to $0.0345$). Temperature scaling has the second smallest average ECE ($0.0427$). However, it is still $23.77\%$ worse than the proposed method. On average, the uncalibrated method and temperature scaling have an $83.08\%$ accuracy, while the proposed method has an $83.51\%$ accuracy. The proposed method increases the accuracy of $11$ out of $16$ tests. 

It is worth noting that temperature scaling increases calibration error of $3$ out of $4$ models on the Kather 5000 dataset, while the proposed method is still able to reduce the calibration error of most cases on the same dataset. The Kather 5000 dataset is a relatively simple and large dataset for its task. The dataset is considered as the MNIST of histology images. It is speculated to have a sufficient amount of training data to train a model end-to-end, with a smaller overfitting effect (i.e., miscalibration). In such a case, since the temperature parameter ($T$) of temperature scaling is learned on only the validation set, it may actually hurt the calibration. However, the proposed method jointly optimizes the accuracy and modal calibration simultaneously, and it can still reduce the calibration error. 
\begin{table}[!tb]
	\centering
	\caption{Expected Calibration Error (ECE) for Each Model}
	\begin{tabular}{|c|c||c|c|c||c|c|}
    \hline
    \multirow{3}{*}{\textbf{Dataset}} & \multirow{3}{*}{\textbf{Model}} & \multicolumn{3}{c||}{\textbf{ECE}} & \multicolumn{2}{c|}{\textbf{Accuracy}} \\
    & & \multicolumn{3}{c||}{(smaller is better)} & \multicolumn{2}{c|}{(larger is better)}\\ \cline{3-7}
    &  & \textbf{\hspace{0.1cm}Unca.\hspace{0.1cm}} & \textbf{\hspace{0.1cm}Temp.\hspace{0.1cm}} &
    \textbf{\hspace{0.2cm}DCA\hspace{0.1cm}} & \textbf{\hspace{0.1cm}Unca.\hspace{0.1cm}}
    & \textbf{\hspace{0.2cm}DCA\hspace{0.1cm}} \\ \hline    
    
    \multirow{4}{*}{RSNA} 
      & AlexNet & $\bf{0.0113}$ & $0.0239$ & $0.0120$ & $0.8376$ & $\bf{0.8488}$\\ \cline{2-7}
      & ResNet & $0.0276$ & $0.0231$ & $\bf{0.0122}$ & $0.8569$ & $\bf{0.8762}$\\ \cline{2-7}
      & DenseNet & $0.0102$ & $0.0814$ & $\bf{0.0077}$ & $0.8502$ & $\bf{0.8543}$\\ \cline{2-7}
      & SqueezeNet & $0.0253$ & $0.0317$ & $\bf{0.0097}$ & $0.8671$ & $\bf{0.8841}$\\ \hline

    \multirow{4}{*}{DDSM} 
      & AlexNet & $0.2164$ & $0.0658$ & $\bf{0.0591}$ & $\bf{0.6766}$ & $0.6291$\\ \cline{2-7}
      & ResNet & $0.1844$ & $\bf{0.0307}$ & $0.0798$ & $\bf{0.7195}$ & $0.6987$\\ \cline{2-7}
      & DenseNet & $0.1798$ & $\bf{0.0337}$ & $0.0754$ & $0.7076$ & $\bf{0.7106}$\\ \cline{2-7}
      & SqueezeNet & $0.2173$ & $\bf{0.0458}$ & $0.0805$ & $\bf{0.6853}$ & $0.6771$\\ \hline
      
    \multirow{4}{*}{Mendeley} 
      & AlexNet & $0.1693$ & $0.0396$ & $\bf{0.0273}$ & $0.8585$ & $\bf{0.8785}$\\ \cline{2-7}
      & ResNet & $0.1475$ & $0.0475$ & $\bf{0.0291}$ & $0.8520$ & $\bf{0.8767}$\\ \cline{2-7}
      & DenseNet & $0.1136$ & $0.0746$ & $\bf{0.0285}$ & $0.8331$ & $\bf{0.8796}$\\ \cline{2-7}
      & SqueezeNet & $0.1871$ & $0.0468$ & $\bf{0.0252}$ & $0.8742$ & $\bf{0.8750}$\\ \hline
    
    \multirow{4}{*}{Kather} & AlexNet & $0.0279$ & $0.0344$ & $\bf{0.0243}$ & $\bf{0.9062}$ & $0.9052$\\ \cline{2-7}
      & ResNet & $\bf{0.0248}$ & $0.0318$ & $0.0304$ & $\bf{0.9355}$ & $0.9229$\\ \cline{2-7}
      & DenseNet & $0.0302$ & $0.0286$ & $\bf{0.0237}$ & $0.9385$ & $\bf{0.9410}$\\ \cline{2-7}
      & SqueezeNet & $0.0372$ & $0.0439$ & $\bf{0.0269}$ & $0.8932$ & $\bf{0.9038}$\\ \hline   
  \multicolumn{2}{|c|}{\textbf{Average}} & $0.1006$ & $0.0427$ & $\bf{0.0345}$ & $0.8308$ & $\bf{0.8351}$ \\ \hline 
  \end{tabular}
  \footnotetext{Footnote}
  \begin{flushleft}
 \end{flushleft}
  \label{table:ece_result}
\end{table}

Table~\ref{table:mce} shows the maximum calibration error (MCE) of the compared models trained using the RSNA, DDSM, and Mendeley datasets. Though there is no clear winner on MCE, the proposed model decreases the average MCE by about $3\%$, while temperature scaling increases the MCE slightly. Table~\ref{table:mce_2} shows the MCE results on the Kather 5000 dataset. Both temperature scaling and the proposed method increase the MCE quite well. According to Guo et al., MCE is very sensitive to the number of bins since it measures the worst cases across all bins, making it an improper metric for small test sets~\cite{guo2017calibration}. On average, temperature scaling has an MCE of $0.3401$ across all the evaluated models, the proposed method has an MCE of $0.3398$, and the uncalibrated method has an MCE of $0.2766$. 

\begin{table}[!tb]
	\centering
	\caption{Maximum Calibration Error (MCE) for Binary Classification Tasks}
	\begin{tabular}{|c|c||c|c|c|}
    \hline
    \multirow{3}{*}{\textbf{Dataset}} & \multirow{3}{*}{\textbf{Model}} & \multicolumn{3}{c|}{\textbf{MCE}} \\
    & & \multicolumn{3}{c|}{(smaller is better)} \\ \cline{3-5}
    &  & \textbf{\hspace{0.2cm}Unca.\hspace{0.1cm}} & \textbf{\hspace{0.1cm}Temp.\hspace{0.1cm}} & \textbf{\hspace{0.2cm}DCA\hspace{0.1cm}} \\ \hline    
    \multirow{4}{*}{RSNA} 
      & AlexNet & $0.0291$ & $0.0366$ & $\bf{0.0230}$ \\ \cline{2-5}
      & ResNet & $0.0484$ & $\bf{0.0325}$ & $0.0399$ \\ \cline{2-5}
      & DenseNet & $0.0335$ & $0.2233$ & $\bf{0.0142}$ \\ \cline{2-5}
      & SqueezeNet & $0.0430$ & $0.0663$ & $\bf{0.0270}$ \\ \hline

    \multirow{4}{*}{DDSM} 
      & AlexNet & $0.2527$ & $\bf{0.1545}$ & $0.1800$ \\ \cline{2-5}
      & ResNet & $0.2897$ & $0.1171$ & $\bf{0.1078}$ \\ \cline{2-5}
      & DenseNet & $0.2403$ & $\bf{0.0941}$ & $0.0959$ \\ \cline{2-5}
      & SqueezeNet & $0.332$ & $0.1631$ & $\bf{0.1586}$ \\ \hline
      
  \multirow{4}{*}{Mendeley} 
      & AlexNet & $0.2454$ & $0.4305$ & $\bf{0.1225}$ \\ \cline{2-5}
      & ResNet & $0.2321$ & $\bf{0.1769}$ & $0.297$ \\ \cline{2-5}
      & DenseNet & $0.2653$ & $\bf{0.2477}$ & $0.4898$ \\ \cline{2-5}
      & SqueezeNet & $0.2521$ & $\bf{0.2451}$ & $0.2507$ \\ \hline
  \multicolumn{2}{|c||}{\textbf{Average}} & $0.2812$ & $0.2817$ & $\bf{0.2737}$ \\ \hline 
  \end{tabular}
  \label{table:mce}
\end{table}

\begin{table}[!tb]
	\centering
	\caption{Maximum Calibration Error for Multi-Class Classification Tasks}
	\begin{tabular}{|c|c||c|c|c|}
    \hline
    \multirow{3}{*}{\textbf{Dataset}} & \multirow{3}{*}{\textbf{Model}} & \multicolumn{3}{c|}{\textbf{MCE}} \\
    & & \multicolumn{3}{c|}{(smaller is better)}\\ \cline{3-5}
    & & \textbf{\hspace{0.2cm}Unca.\hspace{0.1cm}} & \textbf{\hspace{0.1cm}Temp.\hspace{0.1cm}} & \textbf{\hspace{0.2cm}DCA\hspace{0.1cm}} \\ \hline    
    
  \multirow{4}{*}{Kather 5000} & AlexNet & $\bf{0.2570}$ & $0.2974$ & $0.7371$ \\ \cline{2-5}
      & ResNet & $0.3072$ & $0.6379$ & $\bf{0.2513}$ \\ \cline{2-5}
      & DenseNet & $\bf{0.2577}$ & $0.8015$ & $0.4268$ \\ \cline{2-5}
      & SqueezeNet & $\bf{0.2566}$ & $0.3237$ & $0.7371$ \\ \hline
      
  \multicolumn{2}{|c||}{\textbf{Average}} & $\bf{0.2696}$ & $0.5151$ & $0.5381$ \\ \hline 
  \end{tabular}
  \label{table:mce_2}
\end{table}

\subsection{Model Representation Learning}
\label{sec:rec_dist}
As a post-processing method, temperature scaling fixes model miscalibration at the output-level, which does not change the learned representation. However, the proposed method integrates calibration into the network training phase, which may help models learn a better representation.
In this section, we firstly use t-SNE~\cite{maaten2008visualizing} plot to visualize features extracted by temperature scaling and DCA on two datasets. Then, we compare the recovered probability distribution of the two methods on a toy dataset.

\begin{figure}[!tb]
    \centering
    \includegraphics[width=.9\textwidth]{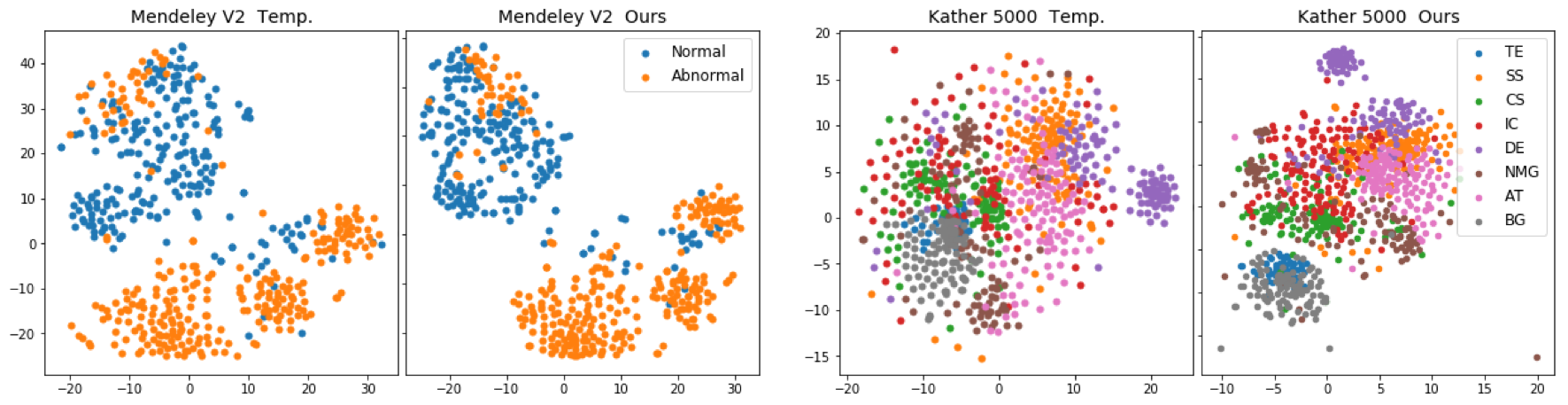}
    \caption{The t-SNE plots of the representations learned using temperature scaling and the proposed method on the Mendeley V2 (left) and Kather 5000 (right) datasets. The samples of temperature scaling are spreading in the feature space (left in each subplot). The samples of the proposed method are densely packed for the same class (right in each subplot).}
    \label{fig:t_sne}
\end{figure}

Figure~\ref{fig:t_sne} shows the t-SNE plots of features extracted by temperature scaling and the proposed method on the Mendeley V2 and Kather 5000 datasets. Each dot represents one data sample in a 2D feature space. To generate these plots, We first extract the high-dimensional feature maps using an AlexNet model trained with either temperature scaling or DCA. We then feed the feature maps to t-SNE that projects the high-dimensional feature maps to 2D feature space. Ideally, the samples from the same class should be close to each other; the samples from different classes should be far from each other. 

The plots reveal that the samples of temperature scaling (left in each subplot) are spreading in the feature space regardless of class labels, while the samples of the same class are densely packed in our method (right in each subplot). Especially for the Kather 5000 dataset, DCA (the rightmost figure) successfully separated the samples of TE (blue) and GB (gray) classes from the rest. But, the same classes of temperature scaling are mixed with others.

\begin{figure}[!tb]
    \centering
    \includegraphics[width=.9\textwidth]{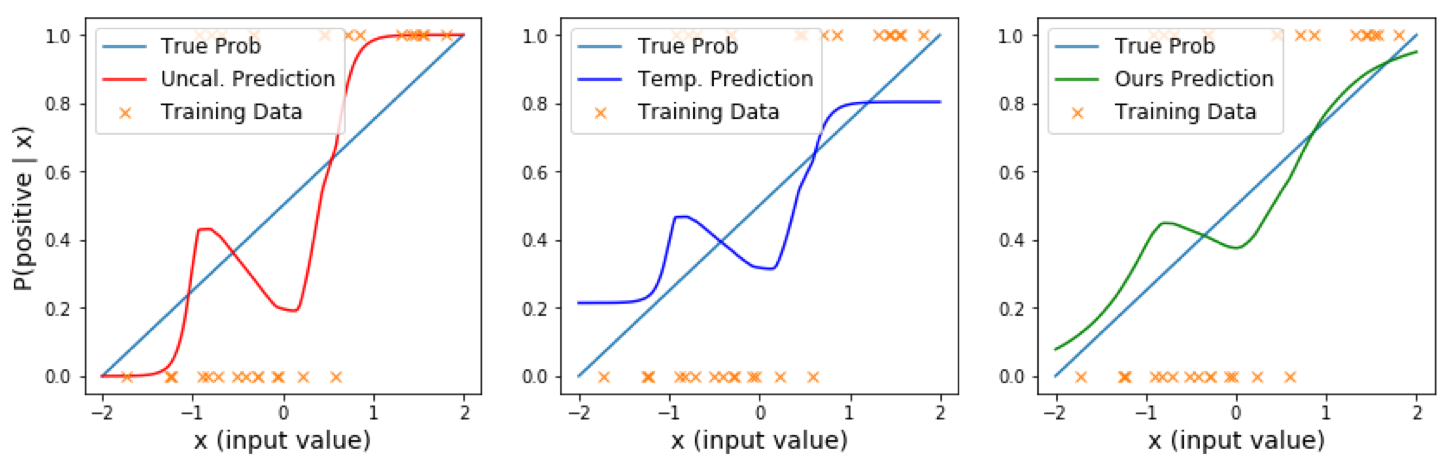}
    \caption{The figure shows the probability distribution that was recovered by the uncalibrated method (left), temperature scaling (middle), and the proposed method (right). The recovered distribution of the uncalibrated model (left) is far from the ground-truth with many overconfident predictions. Temperature scaling (middle) reduces the predicted confidence of the uncalibrated model, but the recovered distribution is still far from the ground-truth. Our method (right) can better recover the true probability.}
    \label{fig:prob}
\end{figure}

Figure~\ref{fig:prob} shows the probability distribution recovered by the uncalibrated method (left), temperature scaling (middle), and the proposed method (right) on a toy dataset. 
For this experiment, we train three simple networks using the uncalibrated method, temperature scaling, and the proposed method, separately. The networks share the same two-layer architecture that takes a one-dimensional input for a binary classification task. We randomly sample a dataset between $-2$ and $2$, and randomly label each sample with either $0$ or $1$. 
The curved line in each figure shows the recovered probability distribution, while the light blue line in each figure shows the ground-truth distribution. 

The figures reveal that the uncalibrated model (left) has many high-probability predictions. For instance, when $-2<x<-1$, the model made many negative predictions with high-probability; when $1<x<2 $, the model made many positive predictions with high-probability. The majority of these predicted probabilities are far from the true probabilities, which indicates the model is overconfident in its predictions and does not capture the true probability distribution well. The temperature scaling method (middle) can relax those extreme predictions by pushing the predicted probabilities close to $0.5$. However, the recovered probability distribution still quite far from the diagonal line. The proposed method (right) can recover the trend of the ground-truth distribution, and most of the predictions are close to the diagonal line. This experiment shows the models trained with DCA may have a strong ability to recover the true probability distribution more accurately. 

\subsection{Hyperparameter Effects}
\begin{figure}[!tb]
    \centering
    \includegraphics[width=.985\textwidth]{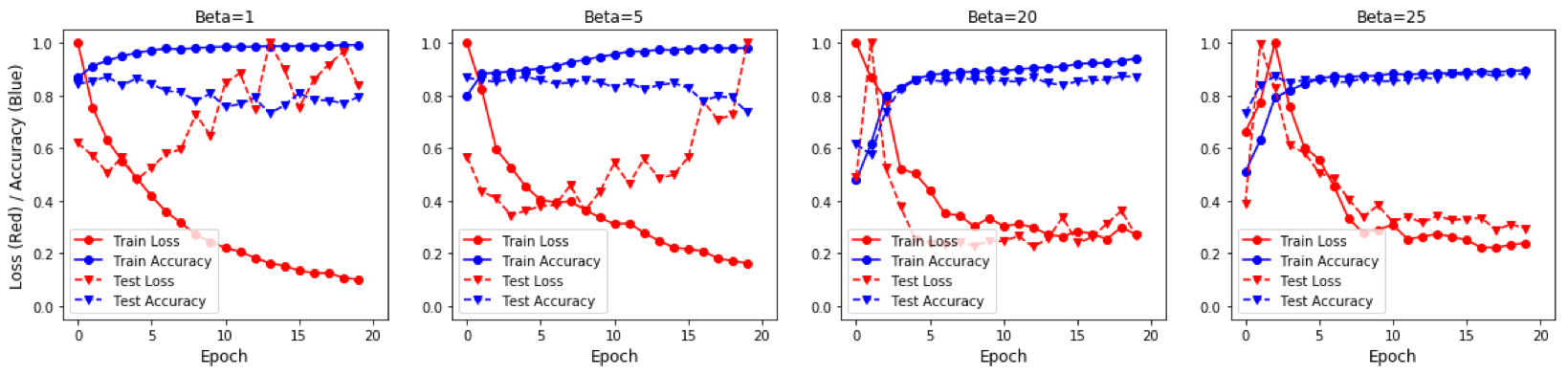}
    \caption{Train/Test accuracy and loss of Mendeley V2 dataset using AlexNet architecture with four different $\beta$ values.}
    \label{fig:mendeley_beta}
\end{figure}

\begin{figure}[!tb]
    \centering
    \includegraphics[width=.985\textwidth]{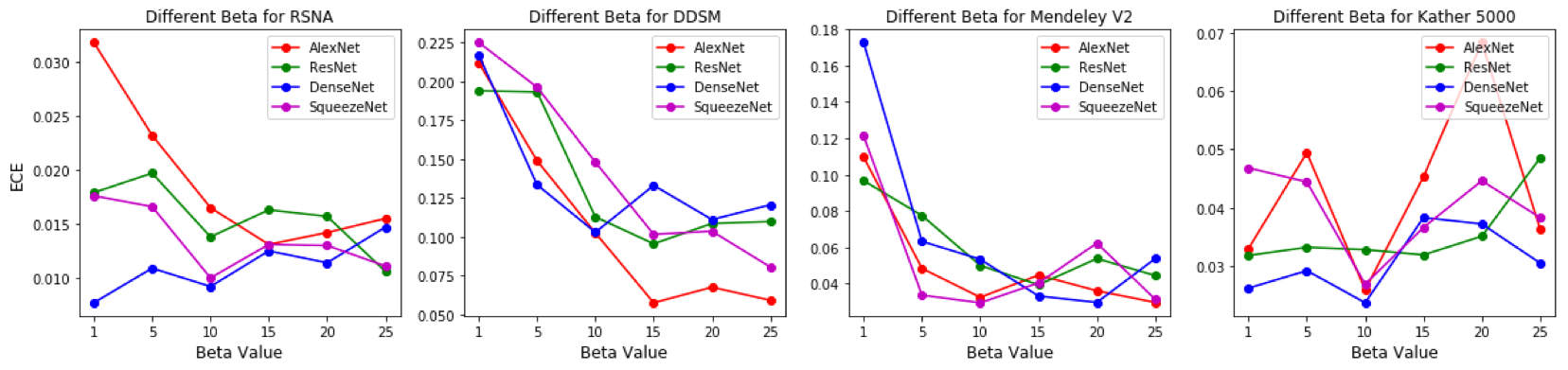}
    \caption{Expected calibration error (ECE) of each dataset with different $\beta$ values.}
    \label{fig:beta_effect}
\end{figure}

One drawback of the proposed method is that the weight scalar $\beta$ needs to be selected for each model. In this section, we show the testing result of the proposed method with different weights ($\beta = [1, 5, 10, 15, 20, 25]$).
Figure~\ref{fig:mendeley_beta} shows the train/test accuracy and loss on the Mendeley V2 dataset using AlexNet architecture with four different $\beta$ values. From the results, we can see that a smaller value such as $1$ or $5$ usually will not be a good choice since they put a smaller penalty to the model when the cross-entropy loss is overfitted. Among our experiments, most of the best results appeared using a weight between $10$ and $25$. Figure~\ref{fig:beta_effect} shows the ECE results of all of the evaluated models with different $\beta$ values. The figure reveals that the ECE result is not very sensitive to the $\beta$ value when $\beta\geq10$, except for the Kather 5000 dataset. In our experiences, we use $\beta$ values between $10$ and $15$ for most of the tasks.

\section{Conclusion}
We proposed a novel approach to neural network calibration that maintains classification accuracy while significantly reducing model calibration error.  We evaluated our approach across various architectures and datasets. Our approach reduces calibration error significantly and comes closer to recovering the true probability than other approaches. The proposed method can be easily integrated into any classification tasks as an auxiliary loss term, thus not requiring an explicit training round for calibration. We believe this simple, fast, and straightforward method can serve as a strong baseline for future researchers.

\bibliographystyle{IEEEtran}  
\bibliography{bibfile}
\newpage
\begin{flushleft}
\Large \textbf{Supplementary Materials}
\end{flushleft}

\renewcommand\thesection{S1}
\section{Datasets}
\label{sec:s2}
\subsection{RSNA}
The RSNA dataset was released for the 2019 RSNA Intracranial Hemorrhage Detection Challenge~\cite{RSNA2019kaggle}. We used the training set of the first stage of the data challenge in this study, which contains $674257$ CT slices of $17079$ patients. The slices were labeled as $7$ classes, normal, intracranial hemorrhage, and five subclasses of intracranial hemorrhage. We used this dataset as a binary classification task (normal/abnormal). The dataset was randomly partitioned into training and testing datasets with a $4:1$ ratio on the patient-level by us.

\subsection{DDSM}
The DDSM dataset contains $2620$ well-labeled cases, including $10480$ digitized screen-film mammography images~\cite{heath2000digital}. The dataset has been almost $20$ years old, which was initially constructed in $1999$. DDSM is the largest publicly available mammography dataset and widely used for developing deep learning models. We chose the Curated Breast Imaging Subset of DDSM (CBIS-DDSM)~\cite{lee2016curated}, which is an updated and standardized version of the original DDSM, for our study. We used the training and testing sets provided by the data provider for training and testing respectively. We used the provided training and testing sets in this study.

\subsection{Mendeley V2}
The Mendeley V2~\cite{kermany2018labeled} dataset contains both of the optical coherence tomography (OCT) images of the retina and pediatric chest X-ray images. We used the pediatric chest X-ray images in this study. The dataset includes $4273$ pneumonia images and $1583$ normal images. We used the provided training and testing sets in this study.

\subsection{Kather 5000}
The Kather 5000~\cite{kather2016multi} dataset contains $5000$ histological images of $150 \times 150$ pixels. Each image belongs to exactly one of eight tissue categories: tumour epithelium, simple stroma, complex stroma, immune cells, debris, normal mucosal glands, adipose tissue, background (no tissue). All images are RGB, $0.495 \mu m$ per pixel, digitized with an Aperio ScanScope (Aperio/Leica biosystems), magnification $20\times$. Histological samples are fully anonymized images of formalin-fixed paraffin-embedded human colorectal adenocarcinomas (primary tumors) from the Institute of Pathology, University Medical Center Mannheim, Heidelberg University, Mannheim, Germany). The dataset was randomly partitioned into training and testing datasets with a $4:1$ ratio by us.

\begin{table}[!h]
	\centering
	\renewcommand\thetable{S1}
	\caption{Datasets used in this study.}
	\begin{tabular}{|c|c|c|c|} 
		\hline
		\textbf{Name}  &  \textbf{\hspace{0.3cm}Modality\hspace{0.3cm}}  &  \textbf{\hspace{0.1cm}\# of Images\hspace{0.1cm}} &  \textbf{\hspace{0.1cm}\# of Classes\hspace{0.1cm}} \\[0.5ex] 
		\hline
		
		\textbf{RSNA} & Head CT & $674257$ & $2$ \\ \hline
		\textbf{DDSM} & Mammography & $10480$ & $2$ \\ \hline
        \textbf{Mendeley V2} & Chest X-ray & $5856$ & $2$  \\ \hline
        \textbf{Kather 5000} & Histological & $5000$ & $8$  \\ \hline
	\end{tabular}
	\label{table:datasets}
\end{table}

\renewcommand\thesection{S2}
\section{CNN Models}
\label{sec:s3}

In this study, we used AlexNet~\cite{krizhevsky2012imagenet}, ResNet-50~\cite{he2016deep}, DenseNet-121~\cite{huang2017densely}, and SqueezeNet 1-1~\cite{iandola2016squeezenet} were as the feature extractor to build our CNN models. More specifically, we firstly pre-trained these four networks on the ImageNet dataset~\cite{deng2009imagenet}. Then, the fully connected (FC) layers and the pooling layer before the FC layers were removed. We froze the parameters of the remaining convolutional (Conv) layers of each network and used them as feature extractors. A shallow CNN classifier was trained on top of each feature extractor. The classifier contained one Conv layer and two FC layers. The Conv layer included convolution, batch normalization~\cite{ioffe2015batch}, leaky ReLU~\cite{xu2015empirical}, and max pooling~\cite{krizhevsky2012imagenet}. Max pooling had a $2\times2$ receptive field with stride $1$. Weighted cross-entropy loss was used in training. Adam optimizer~\cite{kingma2014adam} with a learning rate of $0.0001$ was used as the optimizer. Dropout~\cite{srivastava2014dropout} with a rate of $0.5$ was applied to the FC layers. For the same architecture, all the hyper-parameters were maintained the same among different datasets, except batch sizes. The batch size of AlexNet on DDSM, Mendeley V2, and Kather 5000 datasets was set as $64$, and for RSNA was $512$. The batch sizes were set as half of the AlexNet with the corresponding datasets for the rest of the architectures.

\end{document}